# On Heuristics for Finding Loop Cutsets in Multiply Connected Belief Networks

Jonathan Stillman
*Artificial Intelligence Program*
*General Electric Research and Development Center*
*P.O. Box 8, Schenectady, N.Y. 12301*
*e-mail: stillman@crd.ge.com*

## Abstract

We introduce a new heuristic algorithm for the problem of finding minimum size loop cutsets in multiply connected belief networks. We compare this algorithm to that proposed in [Suermondt and Cooper, 1988]. We provide lower bounds on the performance of these algorithms with respect to one another and with respect to optimal. We demonstrate that no heuristic algorithm for this problem can be guaranteed to produce loop cutsets within a constant difference from optimal. We discuss experimental results based on randomly generated networks, and discuss future work and open questions.

## 1 Introduction

### 1.1 Background and Motivation

One of the central problems in artificial intelligence research is how one can automate reasoning in the presence of uncertain and incomplete knowledge. This type of reasoning is performed regularly by people, but we still lack effective tools for performing such "common-sense" reasoning with computers. Developing such tools is a prerequisite for building advanced expert systems that can cope with the uncertainty and incompleteness that is prevalent in practical domains. There exists a wealth of literature addressing the issues involved (see, for example, [Etherington, 1988] and [Bonissone, 1987]), and a number of ideas for coping with this problem have been suggested. Some of these appear promising, although none of the known approaches can completely alleviate the problem. The use of *belief networks* as a representational paradigm, together with the use of a Bayesian inference mechanism, has recently emerged as a promising approach to handling these issues. Such networks are variously called *belief networks, causal probabilistic networks, influence diagrams,* etc. Belief networks are acyclic, directed graphs in which the nodes represent random variables and the arcs represent dependency relationships that exist between those variables. The basic operation on belief networks is that of calculating and updating the most likely values of certain random variables (representing *hypotheses*) when the values of others is fixed (by evidence generated external to the reasoning system). Some of the most prominent research in this area as it pertains to artificial intelligence can be attributed to Judea Pearl and his colleagues, and is presented in Pearl's recent book [Pearl, 1988]. Bayesian networks have also been used in a number of other areas, among them economics[Wold, 1964], genetics[Wright, 1934], and statistics[Lauritzen and Spiegelhalter, 1988].

### 1.2 Updating in Belief Networks

One of the key problems in developing a practical implementation of a system for reasoning with uncertainty based on Bayesian belief networks is that updating such networks to reflect the impact of new evidence can be computationally costly. Updating belief networks is most simple when the network is *singly connected*, i.e., when there is at most one **undirected** path[1] between any two nodes in the network. Updating such networks can use a relatively efficient *local propagation* algorithm described in [Pearl, 1988]. This is because propagation of evidence in singly connected networks can be done in such a way that information is never multiply accounted for (i.e., the impact of evidence is not fed back to the source of the evidence, and cannot be received along multiple propagation paths). Unfortunately, local propagation techniques are inadequate for networks that contain undirected cycles (we will henceforth refer to such cycles as *loops* in an attempt at differentiating them from directed cycles, which are forbidden by definition of belief networks), called *multiply connected* networks. When the local propagation techniques devised for singly connected networks are used on multiply connected networks, failure may occur in two ways. It is possible that an updating message sent by one node cycles around a loop and causes that node to update again. This repeats indefinitely, causing instability of the network. Even if it does converge, the updated nodes may not have computed the correct posterior probabilities. This is basically due to the fact that certain assumptions of conditional independence that were used by Pearl to derive the local propagation algorithms for singly connected networks fail to hold when the network is multiply connected. Such networks seem to be quite prevalent in practice; thus it is important that effective techniques be developed for handling them. A detailed discussion of Pearl's updating method for singly connected networks is presented in [Pearl, 1988], as is discussion of several approaches to coping with multiply connected networks. Since it is known that the problem of probabilistic inference in belief networks is NP-hard [Cooper, 1990], it is unlikely that exact techniques will be developed that can be guaranteed to yield solutions in an acceptable amount of time. As a result, heuristic techniques need to be explored.

---

[1]Thus, an arc can be traversed in either direction.

### 1.3 Dealing with Multiply Connected Networks

In [Pearl, 1988], Pearl presents three approaches to dealing with the updating problem in multiply connected networks: conditioning, stochastic simulation, and clustering. It is the method of *conditioning* that is of interest to us in this paper. This method relies on identifying a subset of the nodes in the network, elimination of which results in a singly connected network. Such a set of nodes is called a *loop cutset*. Once a loop cutset for a network is identified, the rest of the (now singly connected) network is evaluated for each possible assignment to the random variables represented by the nodes in the cutset, with the results combined by taking a weighted average. This is justified by the rule of conditional probability:

$$p(x|E) = \sum_{c_1,\ldots,c_n} p(x|E, c_1,\ldots,c_n) p(c_1,\ldots,c_n|E)$$

where $E$ is evidence, $x$ is any node in the network, and $c_1,\ldots,c_n$ represents an instantiation of the nodes that form the loop cutset. One important condition must be met by the loop cutset in order to preserve correctness, however: the node that is chosen to cut a loop cannot have multiple parents in the same loop (a good discussion of why this is important can be found in [Suermondt and Cooper, 1988]). Note that since instantiating the loop cutset reduces the belief network to a singly connected network, Pearl's efficient algorithms for such networks can be applied to compute each of the above factors for a given instantiation. The combinatorial difficulty results from the number of instantiations that must be considered. Thus, the complexity of conditioning depends heavily on the size of the loop cutset, being $O(d^c)$, where $d$ is the number of values the random variables can take, and $c$ is the size of the cutset. It is thus important to minimize the size of the loop cutset for a multiply connected network. Unfortunately, the loop cutset minimization problem is easily seen to be NP-hard, using a simple transformation of **Feedback Vertex Set** (see [Garey and Johnson, 1979; Karp, 1972]). Thus it is highly unlikely that one can efficiently compute minimum loop cutsets for large networks, and we must rely on approximation algorithms that yield sub-optimal but hopefully adequate results in many practical cases.

### 1.4 NP-Completeness and Approximation Algorithms

In analyzing the complexity of optimization problems such as the loop cutset problem, it is often useful to examine the corresponding *decision problem* (in this case, the question of whether there exists a loop cutset of size $k$, where $k$ is specified as part of the query). The loop cutset decision problem is *NP-complete*. NP is defined to be the class of languages accepted by a nondeterministic Turing machine in time polynomial in the size of the input string. The NP-complete languages are the "hardest" languages[2] in NP. NP-complete languages share the property that all languages in NP can be transformed into them via some polynomial time transformation. For a thorough discussion of the topic the reader is referred to [Garey and Johnson, 1979]. The fastest known deterministic algorithms for NP-complete problems take time exponential in the problem size in the worst case. It is not known whether this is necessary: one of the central open problems in computer science is whether P = NP. Most researchers believe that P ≠ NP, and that NP-complete problems really do need exponential time to solve. Thus these problems are considered *intractable*, since if P ≠ NP, we cannot hope to correctly solve all instances of them with inputs of nontrivial size.

Knowing that a decision problem is NP-complete does not necessarily suggest that the corresponding optimization problem cannot be approached: sometimes (e.g., the Traveling Salesman Problem) good polynomial approximation algorithms have been devised. Although it is quite difficult in general, it is important to be able to evaluate how well an approximation algorithm can be expected to perform compared to the optimal algorithm. Quite often, algorithms that were purported to work "quite well in practice" behave poorly in general, and only work well on a restricted class of problem instances, which usually goes unidentified.

### 1.5 Results to be Discussed

In the following sections of this paper we will discuss two approximation algorithms for the minimum loop cutset problem. We will discuss an algorithm presented in [Suermondt and Cooper, 1988], introduce our modification of that algorithm, and compare the performance of each of these to each other as well as to the optimal solution In [Suermondt and Cooper, 1988], the authors suggest that their algorithm returns a loop cutset that is "generally small, but that is not guaranteed to be minimal." We show that both their algorithm and ours can perform quite badly with respect to optimal, and furthermore that no polynomial time approximation algorithm for this problem can be guaranteed to return a loop cutset that differs in size from that of the optimal solution by a constant. We discuss empirical results based on implementations of both heuristics and an optimal algorithm, run on random graphs. Finally, we summarize and discuss future work.

---

[2]NP-completeness is often discussed in terms of *decision problems* rather than languages, although the two are interchangeable.

## 2 Loop Cutset Approximation Algorithms

In [Suermondt and Cooper, 1988], a polynomial time heuristic algorithm is provided for the loop cutset problem, together with some empirical analysis. The algorithm, which we will henceforth refer to as $A_1$, consists of two basic parts, which are summarized below.

**Step 1** This step is based on the fact that no node that is part of a singly connected subgraph can break a loop. Each such node is removed from the graph. This is done by iteratively removing each node of degree 1, together with its incident arc. This is repeated until each remaining node has degree greater than 1.

**Step 2** The second step of the algorithm starts by selecting the node of highest degree that has at most one parent, adding that node to the cutset. In the case of ties, the node that can be assigned the largest number of values among those tied is selected.

The algorithm proceeds by repeating steps 1 and 2 above until the remaining graph is singly connected. Modifications of the heuristic are considered that vary in how they weight the relative importance of a candidate node's degree and the number of values it can be assigned.

The heuristic algorithm we have developed, $A_2$, is also a greedy approach. It differs from $A_1$, however, in two important ways. First, the nodes we consider as candidates for Step 2 are a strict superset of those considered by $A_1$: although $A_1$ disallows any node with multiple parents, many such nodes may be viable candidates. In $A_2$, only those nodes that have multiple parents *in the same loop* are disallowed. This difference between the algorithms is shown in Figure 1. Second, we use a more refined scheme for eliminating nodes from the graph that cannot be part of any cutset. To do this, in $A_2$ we augment Step 1 of $A_1$ with a test that checks each remaining node to make sure that it is part of at least one loop. In this way, $A_2$ removes cases such as that shown in Figure 2, where $A_1$ would pick node $v$ if it is of highest degree, even though it cannot be part of any loop, and thus can always be eliminated from consideration. These tests may also allow $A_2$ to identify subgraphs of the original graph that can be processed independently, perhaps decomposing the graph into parts small enough to be processed using an optimal algorithm.

## 3 Performance of the Approximation Algorithms

In this section we discuss bounds on the performance of the two approximation algorithms presented above. We compare the performance of each heuristic to optimal, compare the heuristics to one another, and discuss the possibility of finding *any* suitably good heuristic algorithm for the loop cutset problem. In particular, we have the following theorems:

**Theorem 1** *There exist, for $n > 0$, planar directed acyclic graphs with $O(n)$ vertices and $O(n)$ arcs for which the smallest loop cutset is of size 2, but for which algorithms $A_1$ and $A_2$ return loop cutsets of size $\Omega(\lfloor \frac{n}{4} \rfloor)$.*

**Proof:** We provide a method of constructing graphs, given $n$, for which both of the heuristic algorithms perform as poorly as stated in the theorem. The construction relies on the fact that nodes that have multiple parents in the same loop cannot be chosen. This proof, and the proofs of Theorems 2 and 3 below, share a subgraph that is shown in Figure 3 below. In each case we will add a graph to this graph that prevents the vertex $v$ from being chosen until many others have already been added to the cutset. Figure 3 actually represents a class of graphs, which are constructed by "concatenating" copies of the graph shown in Figure 4, with the rightmost node of one copy being identified with the leftmost node of the next. Note that the set $\{V\}$ forms a loop cutset for each graph constructed in this manner. In order to prevent the greedy algorithms $A_1$ and $A_2$ from choosing this node immediately (and thus forcing them to choose a large number of others), the graph of Figure 3 is modified as shown in Figure 5 below.

In the following discussion, we will use the labels given in Figure 5 to refer to nodes and to how the algorithms choose loop cutsets. Since node $V$ has multiple parents in the same loop, it cannot be chosen. The optimal approach involves choosing one of $\{A, B, C\}$, then choosing $\{V\}$, yielding a loop cutset of size 2. The heuristics, however, choose either node-sets $\{3, 7, 11, 4i + 3, \ldots\}$ together with one of $\{A, B, C\}$, or $\{5, 9, 13, 4i + 5, \ldots\}$ together with $\{V\}$ and one of $\{A, B, C\}$, since all nodes of degree 3 that can break a loop are chosen before any of degree 2, and node $V$ cannot be chosen until a node of degree 2 is chosen. It is easily seen that if there are $n$ nodes in the graph, $\Omega(\lfloor \frac{n}{4} \rfloor)$ are chosen by the heuristics, and the result follows. □

It is interesting to note that the poor performance of the heuristics is not induced by complex graph structures having many more edges than vertices (planar graphs with $n$ vertices cannot have more than $3n - 6$ edges).

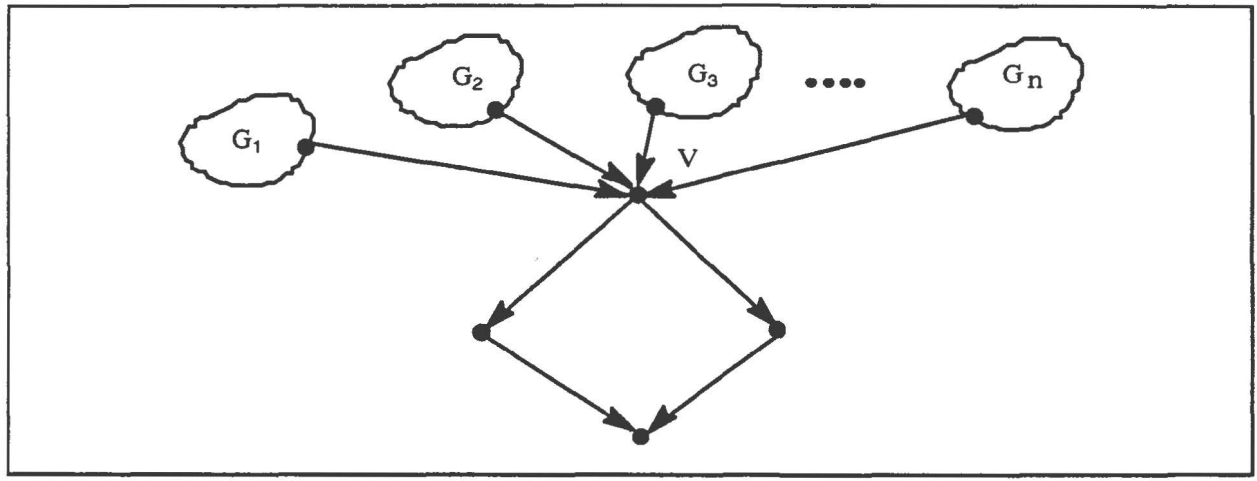

Figure 1: Node V may be chosen by $A_2$ but not by $A_1$.

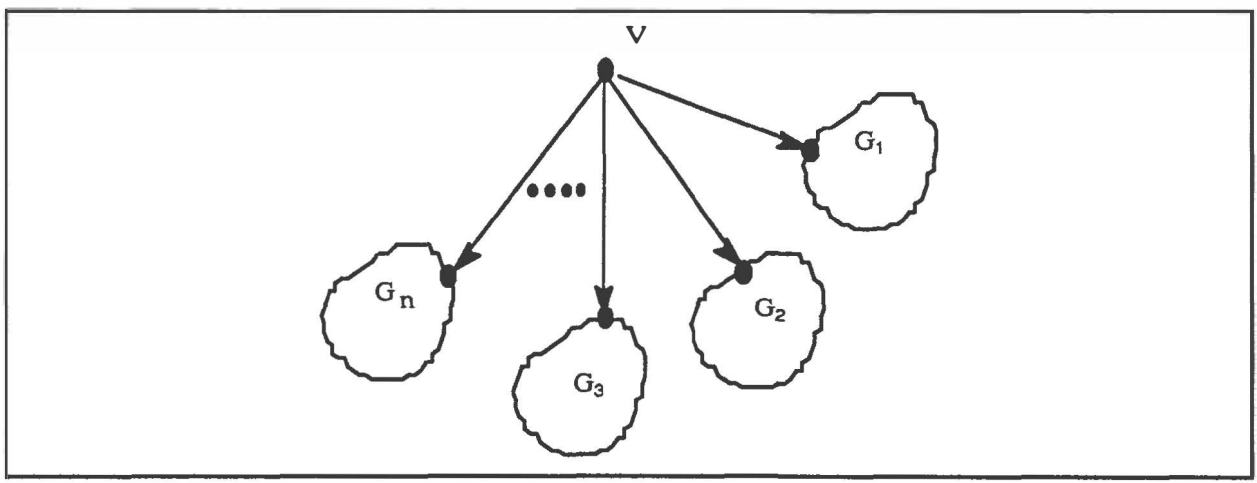

Figure 2: Node V is eliminated by $A_2$ but may be chosen by $A_1$.

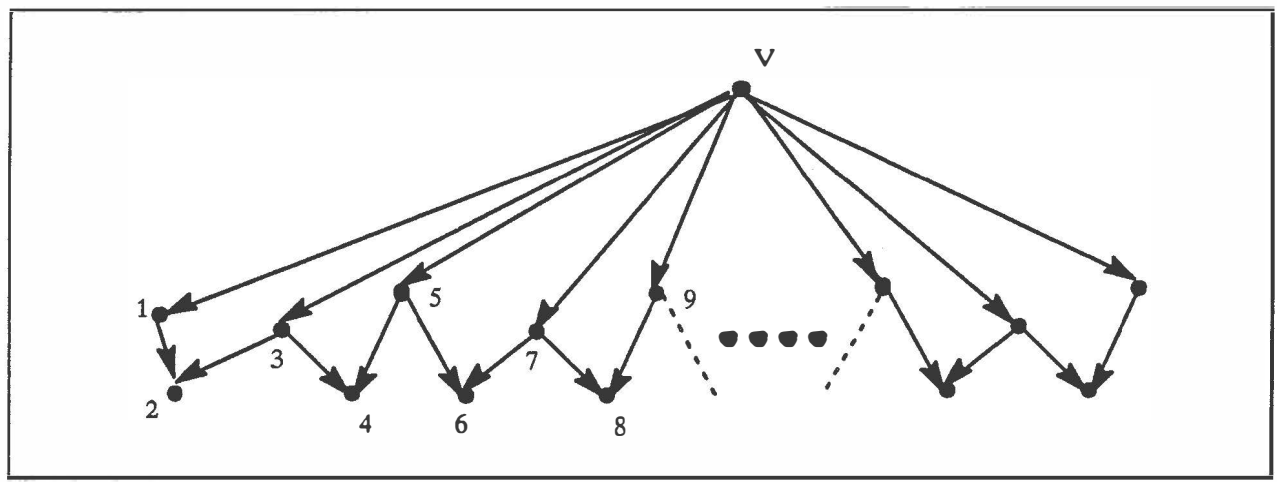

Figure 3: This graph is used in the proofs of Theorems 1, 2, and 3.

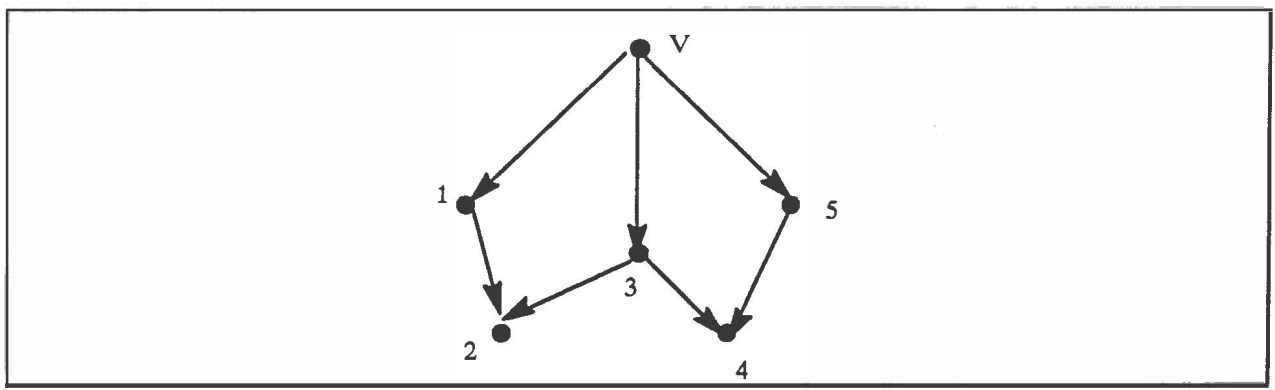

Figure 4: The subgraph that is repeated to make the graph in Figure 3.

**Theorem 2** *There exist, for $n > 0$, planar directed acyclic graphs with $O(n)$ vertices and $O(n)$ arcs for which the smallest loop cutset is of size 1, and for which algorithm $A_2$ performs optimally, but for which algorithm $A_1$ returns a loop cutset of size $\Omega(\lfloor \frac{n}{4} \rfloor)$.*

**Theorem 3** *There exist, for $n > 0$, planar directed acyclic graphs with $O(n)$ vertices and $O(n)$ arcs for which the smallest loop cutset is of size 4, and for which algorithm $A_1$ performs within 2 nodes of optimal, but for which algorithm $A_2$ returns a loop cutset of size $\Omega(\lfloor \frac{n}{4} \rfloor)$.*

Due to space limitations, we omit the proofs of Theorems 2 and 3. They rely on similar constructions to that developed for the proof of Theorem 1, but are a bit more complex. The proof of Theorem 4 is also omitted due to space considerations. Complete proofs of all theorems presented appear in the full version of this paper [Stillman, 1989].

Considering the negative results presented above, it is natural to ask how well one can expect an arbitrary approximation algorithm to perform on this problem. The next theorem is a partial answer to that question.

**Theorem 4** *If $P \neq NP$, then no polynomial approximation algorithm can be guaranteed to find, given an arbitrary directed acyclic graph, a loop cutset whose size differs from that of the smallest loop cutset for the graph by a constant.*

Thus it is not the case that the algorithms $A_1$ and $A_2$ are simply weak heuristics: under the (reasonable) assumption that $P \neq NP$, *no* approximation algorithm is always going to find a loop cutset within a constant difference from optimal.

Since each extra node in the loop cutset at least doubles the computational cost of belief updating, it is important to determine how these heuristics perform in practice. We explore this in the next section. A more detailed discussion can be found in [Stillman, 1989].

## 4 Experimental Results

We have implemented the two approximation algorithms discussed above, and have run them on a number of belief networks. We have also implemented an algorithm to find a loop cutset of optimal size[3]. In this section we present the results of our comparison of the two approximation algorithms with one another, and, where feasible, how each compared with the optimal. These tests were run on networks that were generated randomly in two ways. We outline below the network generation approaches we used. These approaches seem to produce networks quite similar to those that arise in practice; samples of the graphs generated can be found in [Stillman, 1989].

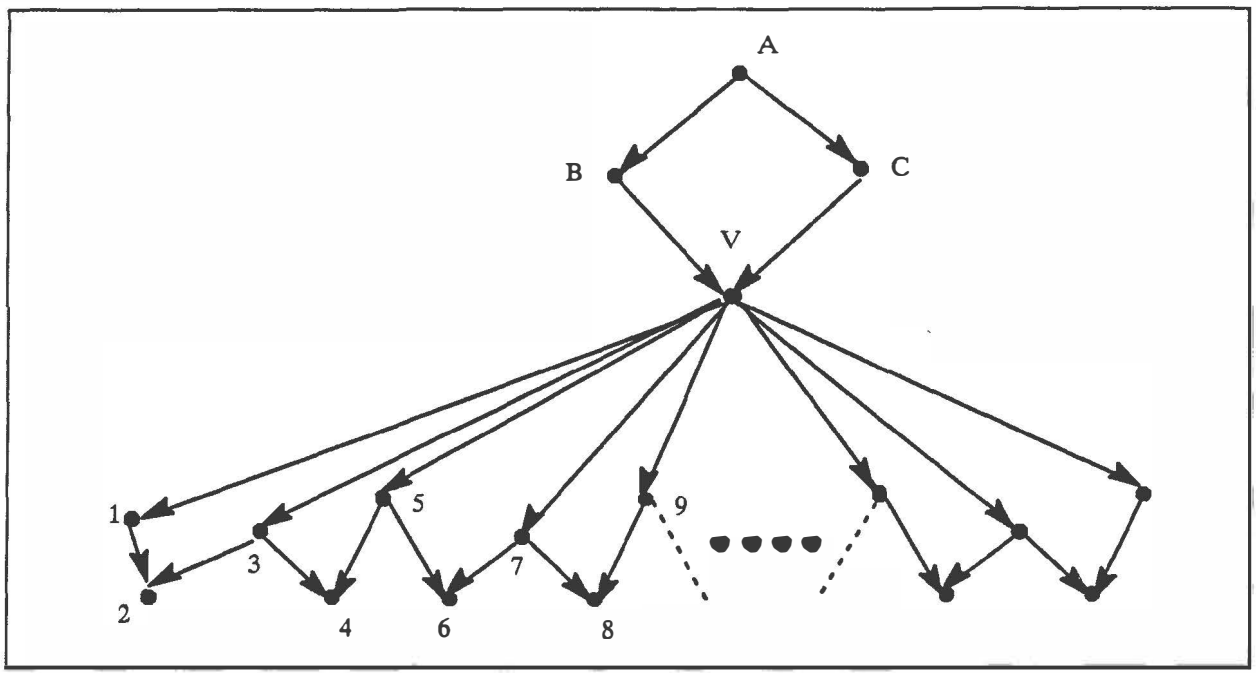

Figure 5: A graph for which $A_1$ and $A_2$ perform poorly.

### 4.1 Generating Test Graphs

The first algorithm ($G_1$) for generating random graphs requires the user to specify the number of nodes desired and the probability of an arc from one arbitrary node to another. The specified number of nodes are created, and for each pair, with the probability specified, an arc is added. Acyclicity (in the directed sense) is preserved by numbering the nodes and directing all added arcs from the lower numbered node to the higher.

The second algorithm ($G_2$) was similar to that described in [Suermondt and Cooper, 1988]. The user specifies the number of nodes and the number of edges. The algorithm proceeds to create all possible arcs in the graph (maintaining acyclicity as above). Arcs are deleted randomly until the desired number of arcs remains. As an option that can be specified by the user, the algorithm can be instructed to ensure that the resulting graph remains connected as each arc is deleted, choosing another if deletion would disconnect the graph.

### 4.2 Test Results

We ran both $A_1$ and $A_2$ on several thousand graphs generated using algorithms $G_1$ and $G_2$. Where feasible, we also compared the heuristics with an optimal solution. The tables below summarize the experiments. In most cases the two approximation algorithms performed comparably. Where there was discrimination, our algorithm almost always performed better than $A_1$ (approximately 95% of the time). Each row in each of the tables summarizes the performance of the heuristic algorithms on 100 graphs of the given size (in the case of $G_1$ this is a combination of number of nodes and probability of an edge, in the case of $G_2$, it is measured by the number of nodes and the number of edges).

Figure 6 contains charts that summarize the results of a number of tests on graphs whose complexity allowed comparison of the heuristic algorithms with optimal. Each chart compares the performance of the three algorithms on 100 graphs of a specified class. If all three returned the same size cutset, no entry was made. Thus, the charts show how each algorithm performed on a given graph when at least one of the heuristics performed sub-optimally. The size of the smallest cutset is also shown in each case.

## 5 Conclusions and Future Work

We have explored several aspects of the loop cutset problem for multiply connected belief networks, modifying a known heuristic algorithm in a way that shows improved performance in empirical studies. We have shown, however, that the known heuristics may perform arbitrarily poorly, even when the graphs under consideration are relatively sparse, being planar directed acyclic graphs. Furthermore, we have shown that the two heuristics we studied are incomparable (i.e., each may perform arbitrarily poorly with respect to the other on a given problem instance), and that no heuristic algorithm for this problem can be guaranteed to return a loop cutset that differs from optimal by a constant on all graphs.

---
[3]We also experimented with a random heuristic, which simply picked nodes from the graph at random until all loops were broken. This heuristic performed poorly in general.

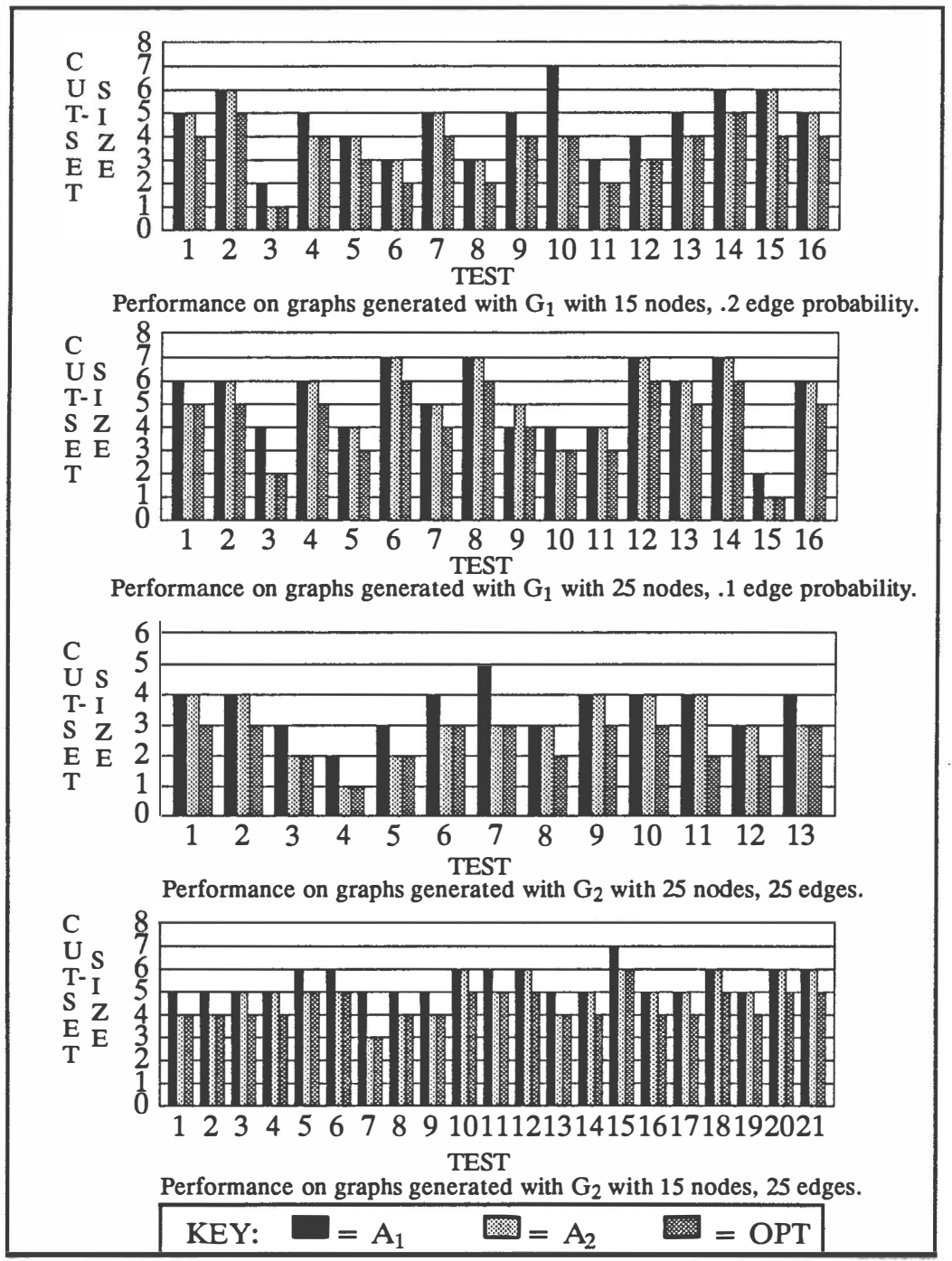

Figure 6: Summary of performance where measurable against optimal.

| #nodes | arc-prob. | $A_1 = A_2$ | $A_1 < A_2$ | $A_1 > A_2$ |
|---|---|---|---|---|
| 15 | .2 | 94 | 0 | 6 |
| 20 | .1 | 95 | 0 | 5 |
| 25 | .1 | 95 | 1 | 4 |
| 50 | .05 | 84 | 3 | 13 |
| 50 | .1 | 83 | 0 | 17 |
| 100 | .02 | 84 | 1 | 15 |

Table 1: Performance of the heuristic algorithms on graphs generated using $G_1$.

| #nodes | #arcs | $A_1 = A_2$ | $A_1 < A_2$ | $A_1 > A_2$ |
|---|---|---|---|---|
| 25 | 25 | 96 | 0 | 4 |
| 25 | 50 | 89 | 0 | 11 |
| 25 | 75 | 94 | 0 | 6 |
| 50 | 50 | 93 | 2 | 5 |
| 50 | 100 | 85 | 0 | 15 |
| 100 | 100 | 82 | 1 | 17 |

Table 2: Performance of the heuristic algorithms on graphs generated using $G_2$.

We leave open the question of whether there exists a heuristic algorithm for this problem can be guaranteed to return a loop cutset that differs from optimal by a constant multiplicative factor on all graphs, or whether there exists any fixed constant for which this is true. There are very few such results appearing in the literature; the interested reader is referred to [Garey and Johnson, 1979] for details. In addition, further study of the empirical performance of our heuristic is needed. The heuristic will be incorporated into an implementation of a Bayesian reasoning tool under development at General Electric. We will test the applicability of our method to problems in several areas of interest, as well as comparing it with other approaches to coping with loops in Bayesian networks, such as those discussed in [Lauritzen and Spiegelhalter, 1988] and in [Pearl, 1988].